\documentclass{article}
\usepackage{ijcai22}

\usepackage{times}
\usepackage{soul}
\usepackage{url}
\usepackage[hidelinks]{hyperref}
\usepackage[utf8]{inputenc}
\usepackage[small]{caption}
\usepackage{graphicx}
\usepackage{amsmath}
\usepackage{amsthm}
\usepackage{booktabs}
\usepackage[ruled]{algorithm2e}
\usepackage{graphicx}

\usepackage{tikz}
\usepackage{comment}
\usepackage{amsmath,amssymb} % define this before the line numbering.
\usepackage{color}
\usepackage{xcolor}

\usepackage{graphicx}
\usepackage{amsmath}
\usepackage{amssymb}
\usepackage{booktabs}
\usepackage{amsmath}
\usepackage{tabularx}

\title{RSIR Transformer: Hierarchical Vision Transformer using Random Sampling Windows and Important Region Windows}

\author{
Zhemin Zhang$^1$
\and
Xun Gong$^1$
\affiliations
$^1$Southwest Jiaotong University
\emails
zheminzhang@my.swjtu.edu.cn
}

\begin{document}

\maketitle

\begin{abstract}
Recently, Transformers have shown promising performance in various vision tasks. However, the high costs of global self-attention remain challenging for Transformers, especially for high-resolution vision tasks. Local self-attention runs attention computation within a limited region for the sake of efficiency, resulting in insufficient context modeling as their receptive fields are small. In this work, we introduce two new attention modules to enhance the global modeling capability of the hierarchical vision transformer, namely, random sampling windows (RS-Win) and important region windows (IR-Win). Specifically, RS-Win sample random image patches to compose the window, following a uniform distribution, i.e., the patches in RS-Win can come from any position in the image. IR-Win composes the window according to the weights of the image patches in the attention map. Notably, RS-Win is able to capture global information throughout the entire model, even in earlier, high-resolution stages. IR-Win enables the self-attention module to focus on important regions of the image and capture more informative features. Incorporated with these designs, RSIR-Win Transformer demonstrates competitive performance on common vision tasks.
\end{abstract}

\section{Introduction}

Modeling in computer vision has long been dominated by convolutional neural networks (CNNs). Recently, transformer models in the field of natural language processing (NLP) \cite{DBLP:journals/corr/abs-1810-04805,NIPS2017-3f5ee243,10.1145/3437963.3441667} have attracted great interest from computer vision (CV) researchers. The Vision Transformer (ViT) \cite{DBLP:journals/corr/abs-2010-11929} model and its variants have gained state-of-the-art results on many core vision tasks \cite{Zhao2020CVPR,pmlr-v139-touvron21a}. The original ViT, inherited from NLP, ﬁrst splits an input image into patches, while equipped with a trainable class (CLS) token that is appended to the input patch tokens. Then, patches are treated in the same way as tokens in NLP applications, using self-attention layers for global information communication, and finally using the output CLS token for prediction. Recent work \cite{DBLP:journals/corr/abs-2010-11929,Liu-2021-ICCV} shows that ViT outperforms state-of-the-art convolutional networks \cite{Huang-2018-CVPR} on large-scale datasets. However, when trained on smaller datasets, ViT usually underperforms its counterparts based on convolutional layers.

The original ViT lacks inductive bias, such as locality and translation equivariance, which leads to overﬁtting and data inefficient usage. To improve data efficiency, numerous eﬀorts have studied how to introduce the locality of the CNN model into the ViT to improve its scalability \cite{NEURIPS2021-4e0928de,DBLP:journals/corr/abs-2107-00641}. These methods typically re-introduce hierarchical architectures to compensate for the loss of non-locality, such as the Swin Transformer \cite{Liu-2021-ICCV}. 

Local self-attention and hierarchical ViT (LSAH-ViT) has been demonstrated to solve data inefﬁciency and alleviate model overfitting. However, LSAH-ViT uses window-based attention at shallow layers, losing the non-locality of original ViT, which leads to LSAH-ViT having limited model capacity and henceforth scales unfavorably on larger datasets such as ImageNet-21K \cite{NEURIPS2021-20568692}. To bridge the connection between windows, previous LSAH-ViT works propose specialized designs such as the “haloing operation” \cite{Vaswani-2021-CVPR} and “shifted window” \cite{Liu-2021-ICCV}. These approaches often need complex architectures, and their receptive field is increased quite slowly and requires stacking many blocks to achieve global self-attention.

\begin{figure*}[t]
\centering
\includegraphics[width=0.9\linewidth]{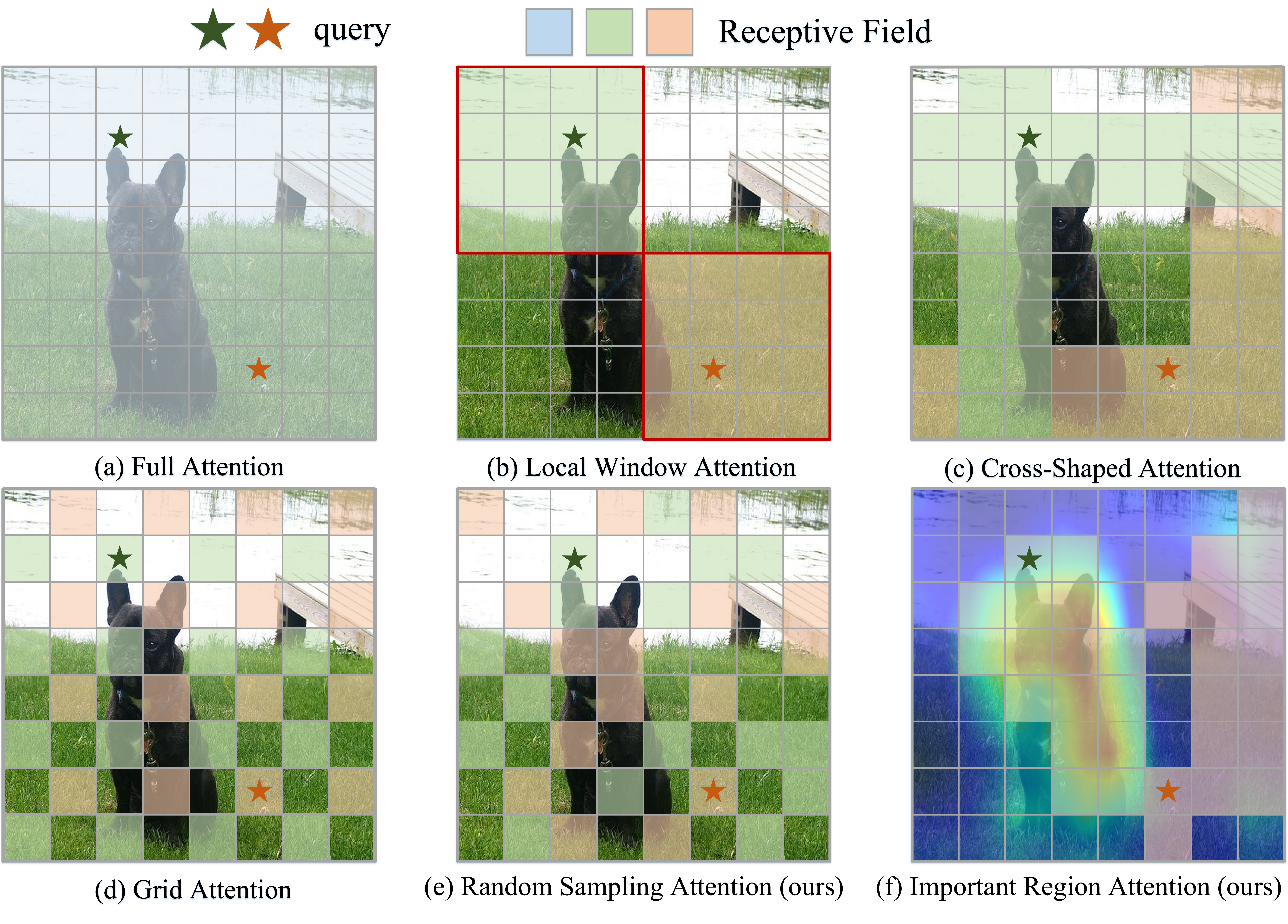} % Reduce the figure size so that it is slightly narrower than the column.
\caption{Full attention and its sparse variants. (a) Full attention is a global operation that is computationally expensive and requires a lot of memory. (b) Swin Transformer uses partitioned window attention. (c) Cross-shaped attention parallel computes the self-attention of horizontal and vertical stripes that form a cross-shaped window. (d) Grid attention attends globally to patches in a sparse, uniform grid overlaid on the entire 2D space. (e) Random sampling attention sample random patches compose a window and perform self-attention within the window. (f) Important region attention composes the window according to the weights of the image patches in the attention map. The same colors are spatially mixed by the self-attention operation.}
\label{DifferentAttentionMechanisms-flabel}
\end{figure*}

In this paper, we propose two novel types of Transformer modules, called random sampling windows (RS-Win) and important region windows (IR-Win). RS-Win sample random image patches to compose the window, following a uniform distribution, i.e., the patches in RS-Win can come from any position in the image, shown in Figure \ref{DifferentAttentionMechanisms-flabel}(e). RS-Win can perform both local and global spatial interactions in a single block. Compared to the previously handcrafted sparse attention, RS-Win gives greater flexibility. The number of patches randomly sampled in each window of RS-Win is fixed, and computing self-attention locally within windows, thus the complexity becomes linear to image size.

To make the model have not only "randomness" but also " determinism", we use IR-Win to make the model focus on important regions of the feature map. IR-Win first uses a method similar to that in CBAM \cite{Woo-2018-ECCV} to infer the attention map of the input feature map, which marks which regions of the image are important for the recognition task. With the attention map determined, the next step is to sample according to the attention map. That is, patches with high attention weights compose a window, and similarly, patches with low weights compose other windows, as shown in Figure \ref{DifferentAttentionMechanisms-flabel}(f). Here, we can simply view RS-Win as "exploration", whose purpose is to freely "explore" which image patches are important in the image. The purpose of IR-Win is to integrate important image patches obtained from "exploration" into the same window for information interaction. By combining RS-Win and IR-Win, each local window can flexibly explore the entire image space and receive data-dependent local information, facilitating learning more complex relations.

Based on the proposed RS-Win and IR-Win self-attention, we design a general vision transformer backbone with a hierarchical architecture, named RSIR Transformer. Our tiny variant RSIR-T achieves 83.9\% Top-1 accuracy on ImageNet-1K without any extra training data.

\section{Related Work}

Transformers were proposed by Vaswani et al. \cite{NIPS2017-3f5ee243} for machine translation, and have since become the state-of-the-art method in many NLP tasks. Recently, ViT \cite{DBLP:journals/corr/abs-2010-11929} demonstrates that pure Transformer-based architectures can also achieve very competitive results. One challenge for vision transformer-based models is data efficiency. Although ViT \cite{DBLP:journals/corr/abs-2010-11929} can perform better than convolutional networks with hundreds of millions of images for pre-training, such a data requirement is difficult to meet in many cases. 

To improve data efficiency, many recent works have focused on introducing the locality and hierarchical structure of convolutional neural networks into ViT, proposing a series of local and hierarchical ViT. The Swin Transformer \cite{Liu-2021-ICCV} pays attention on shifted windows in a hierarchical architecture. Nested ViT \cite{zhang2022nested} proposes a block aggregation module, which can more easily achieve cross-block non-local information communication. Focal ViT \cite{DBLP:journals/corr/abs-2107-00641} presents focal self-attention, each token attends its closest surrounding tokens at ﬁne granularity and the tokens far away at coarse granularity, which can effectively capture both short- and long-range visual dependencies.

Based on the local window, a series of local self-attentions with different shapes are proposed in subsequent work. Axial self-attention \cite{DBLP:journals/corr/abs-1912-12180} and criss-cross attention \cite{Huang-2019-ICCV} achieve longer-range dependencies in horizontal and vertical directions respectively by performing self-attention in each single row or column of the feature map. CSWin \cite{Dong-2022-CVPR} proposed a cross-shaped window self-attention region, including multiple rows and columns. MaxViT \cite{10.1007/978-3-031-20053-3-27} proposed grid attention, the grid attention module attends globally to pixels in a sparse, uniform grid overlaid on the entire 2D space. The performance of the above attention mechanisms are either limited by the restricted window size or has a high computation cost, which cannot achieve a better trade-off between computation cost and global-local interaction.

This paper proposes a new hierarchical vision Transformer backbone by introducing RS-Win and IR-Win self-attention. MaxViT \cite{10.1007/978-3-031-20053-3-27} is the most related works with our RSIR Transformer. Compared to it, RSIR Transformer not only gives greater flexibility for global information communication, but also has data-dependent local information communication.

\begin{algorithm}[tb]
    \caption{\small{ Pseudocode of RS-Win in a PyTorch-like style.}}
    \label{alg:RS-Win}
    %\textcolor[rgb]{0.133, 0.757, 0.133}{\# \textbf{Input:} x.shape=B, L, C (batch, length, dim)}\\  feature maps or
    \KwIn{ x.shape=B, L, C (batch, length, dim)}
    \textcolor[rgb]{0.133, 0.757, 0.133}{\# sample\_map from uniform distribution}\\
    sample\_map = torch.rand(B, L, device=x.device)\\
    
    \textcolor[rgb]{0.133, 0.757, 0.133}{\# Sort sample\_map for each token sequence}\\
    ids\_shuffle = torch.argsort(sample\_map, dim=1)\\
    \textcolor[rgb]{0.133, 0.757, 0.133}{\# ids\_restore for sequence restore}\\
    ids\_restore = torch.argsort(ids\_shuffle, dim=1)\\
    \textcolor[rgb]{0.133, 0.757, 0.133}{\# Shuffle token sequence by ids\_shuffle}\\
    x\_shuffe = torch.gather(x, dim=1, index=ids\_shuffle.unsqueeze(-1).repeat(1, 1, C))\\
    
    \textcolor[rgb]{0.133, 0.757, 0.133}{\#  Partition windows}\\
    x\_windows = window\_partition(x\_shuffe, window\_size)\\
    \textcolor[rgb]{0.133, 0.757, 0.133}{\#  Windows self-attention}\\
    attn\_windows = self.attn(x\_windows)\\
    
    \textcolor[rgb]{0.133, 0.757, 0.133}{\#  Reverse windows}\\
    x = window\_reverse(attn\_windows, window\_size, H, W)\\
    \textcolor[rgb]{0.133, 0.757, 0.133}{\#  Restore the token sequence}\\
    x\_restore = torch.gather(x, dim=1, index=ids\_restore.unsqueeze(-1).repeat(1, 1, C))\\
  	
\end{algorithm}

\section{Method}

\begin{figure}[h]
\centering
\includegraphics[width=0.95\linewidth]{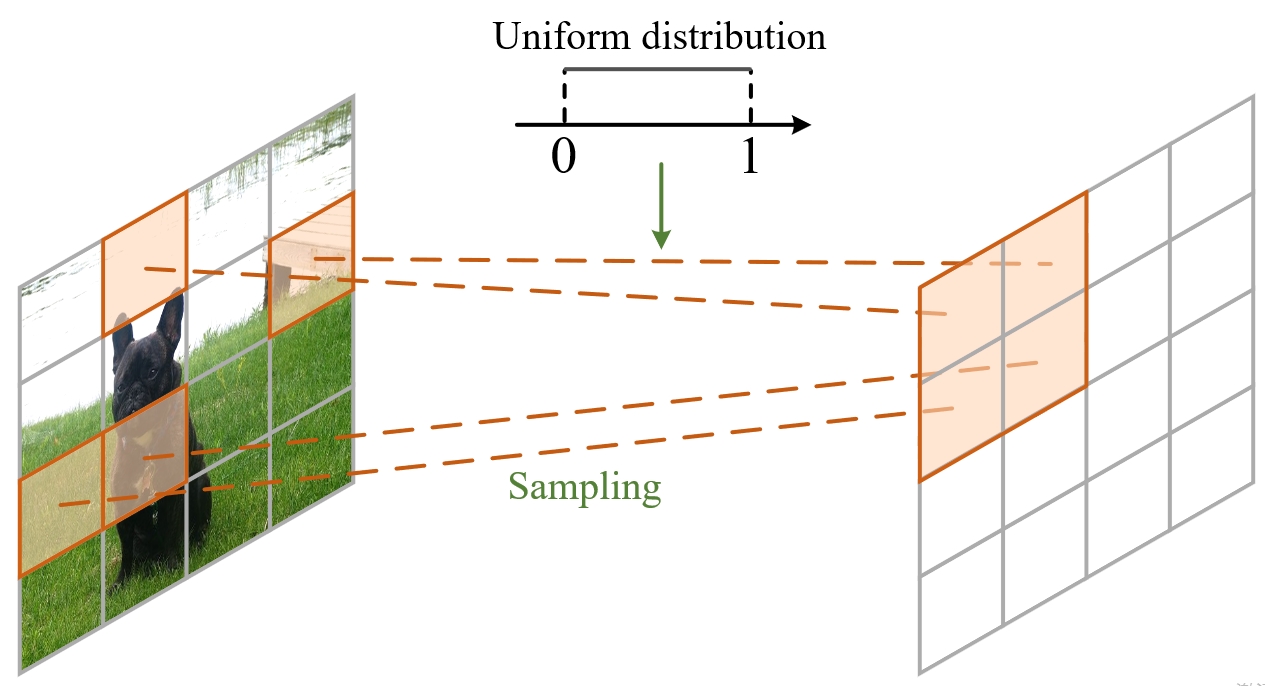} % Reduce the figure size so that it is slightly narrower than the column.
\caption{RS-Win self-attention}
\label{RS-Win-flabel}
\end{figure}

\subsection{Random Sampling Windows Self-Attention}

LSAH-ViT uses window-based attention at shallow layers, lacking the non-locality of original ViT, which leads to LSAH-ViT having limited model capacity and henceforth scales unfavorably on larger datasets. Existing works use specialized designs, such as the “haloing operation” \cite{Vaswani-2021-CVPR} and “shifted window” \cite{Liu-2021-ICCV}, to communicate information between windows. These approaches often need complex architectures, and their receptive field is increased quite slowly and requires stacking many blocks to achieve global self-attention. For capturing dependencies varied from short-range to long-range, we propose RS-Win self-attention. Compared to the previously handcrafted sparse attention, RS-Win gives greater flexibility. 

RS-Win sample random image patches to compose the window, following a uniform distribution, i.e., the patches in RS-Win can come from any position in the image, shown in Figure \ref{RS-Win-flabel}. The RS-Win algorithm is summarized with Pytorch-like pseudo code in Algorithm \ref{alg:RS-Win}.

\begin{figure}[h]
\centering
\includegraphics[width=0.95\linewidth]{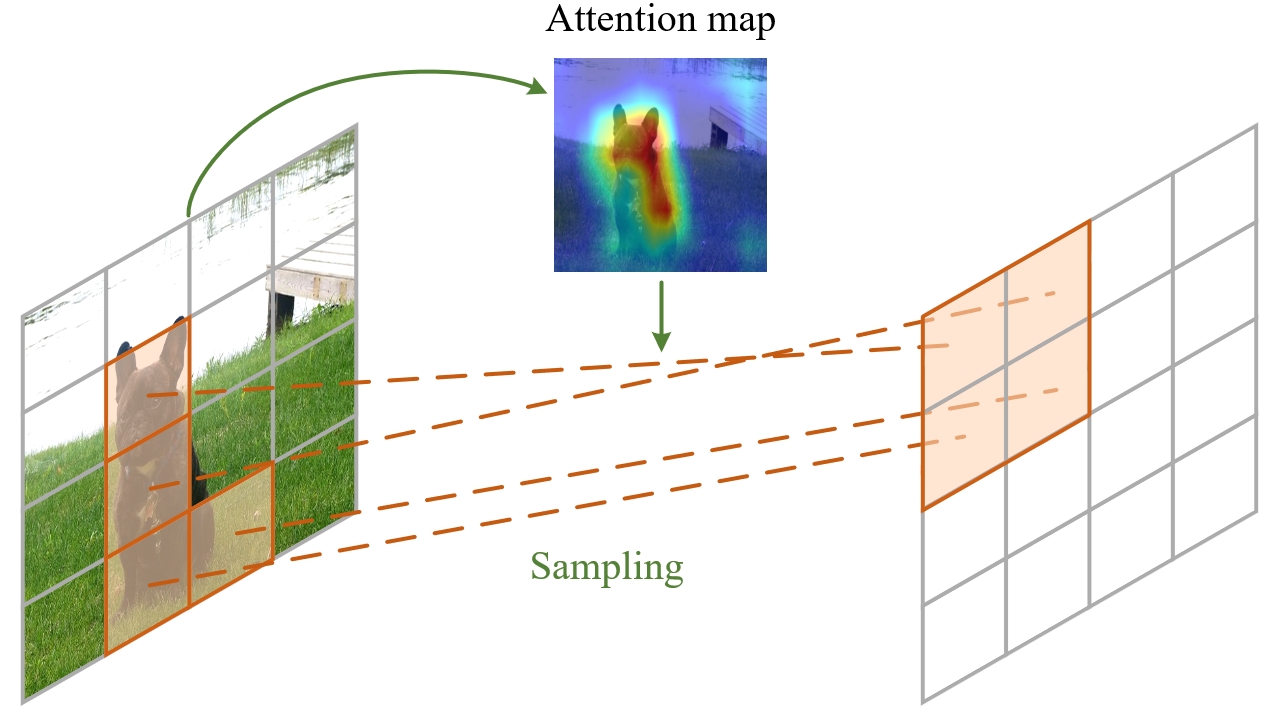} % Reduce the figure size so that it is slightly narrower than the column.
\caption{IR-Win self-attention}
\label{IR-Win-flabel}
\end{figure}

\begin{figure*}[t]
\centering
\includegraphics[width=1.0\linewidth]{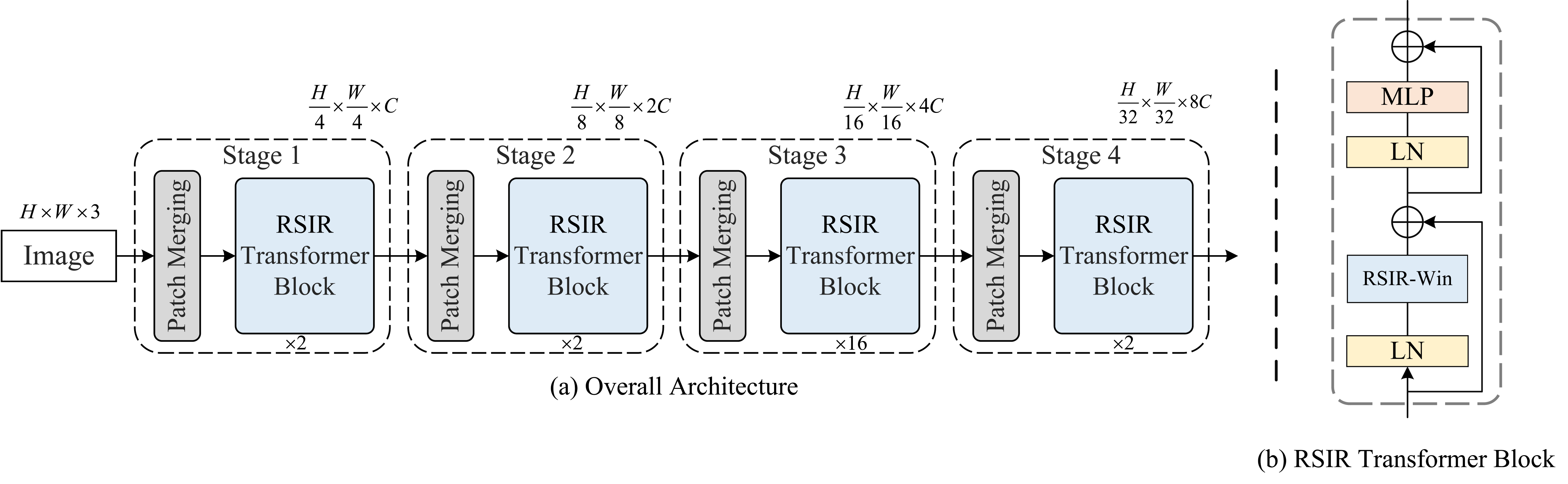} % Reduce the figure size so that it is slightly narrower than the column.
\caption{Left: The overall architecture of our RSIR Transformer, Right: The composition of each block. }
\label{OverallArchitecture-flabel}
\end{figure*}

\begin{algorithm}[tb]
    \caption{\small{ Pseudocode of IR-Win in a PyTorch-like style.}}
    \label{alg:IR-Win}
    %\textcolor[rgb]{0.133, 0.757, 0.133}{\# \textbf{Input:} x.shape=B, L, C (batch, length, dim)}\\  feature maps or
    \KwIn{ x.shape=B, L, C (batch, length, dim)}
    \textcolor[rgb]{0.133, 0.757, 0.133}{\# sample\_map from attention map}\\
    with torch.no\_grad():\\
    \hspace*{1em}  sample\_map = torch.mean(x,dim=2)\\
    
    \textcolor[rgb]{0.133, 0.757, 0.133}{\# Sort sample\_map for each token sequence}\\
    ids\_shuffle = torch.argsort(sample\_map, dim=1)\\
    \textcolor[rgb]{0.133, 0.757, 0.133}{\# ids\_restore for sequence restore}\\
    ids\_restore = torch.argsort(ids\_shuffle, dim=1)\\
    \textcolor[rgb]{0.133, 0.757, 0.133}{\# Shuffle token sequence by ids\_shuffle}\\
    x\_shuffe = torch.gather(x, dim=1, index=ids\_shuffle.unsqueeze(-1).repeat(1, 1, C))\\
    
    \textcolor[rgb]{0.133, 0.757, 0.133}{\#  Partition windows}\\
    x\_windows = window\_partition(x\_shuffe, window\_size)\\
    \textcolor[rgb]{0.133, 0.757, 0.133}{\#  Windows self-attention}\\
    attn\_windows = self.attn(x\_windows)\\
    
    \textcolor[rgb]{0.133, 0.757, 0.133}{\#  Reverse windows}\\
    x = window\_reverse(attn\_windows, window\_size, H, W)\\
    \textcolor[rgb]{0.133, 0.757, 0.133}{\#  Restore the token sequence}\\
    x\_restore = torch.gather(x, dim=1, index=ids\_restore.unsqueeze(-1).repeat(1, 1, C))\\
  	
\end{algorithm}

\subsection{Important Region Windows Self-Attention}

To make the model have not only "randomness" but also " determinism", we use IR-Win to make the model focus on important regions of the feature map. IR-Win first uses a method similar to that in CBAM \cite{Woo-2018-ECCV} to infer the attention map of the input feature map, which marks which regions of the image are important for the recognition task. With the attention map determined, the next step is to sample according to the attention map. That is, patches with high attention weights compose a window, and similarly, patches with low weights compose other windows, as shown in Figure \ref{IR-Win-flabel}. IR-Win enables local information communication to become data-dependent, thereby facilitating the model's ability to distinguish redundant information. The IR-Win algorithm is summarized with Pytorch-like pseudo code in Algorithm \ref{alg:IR-Win}.

\begin{figure}[t]
  \centering
   \includegraphics[width=1.0\linewidth]{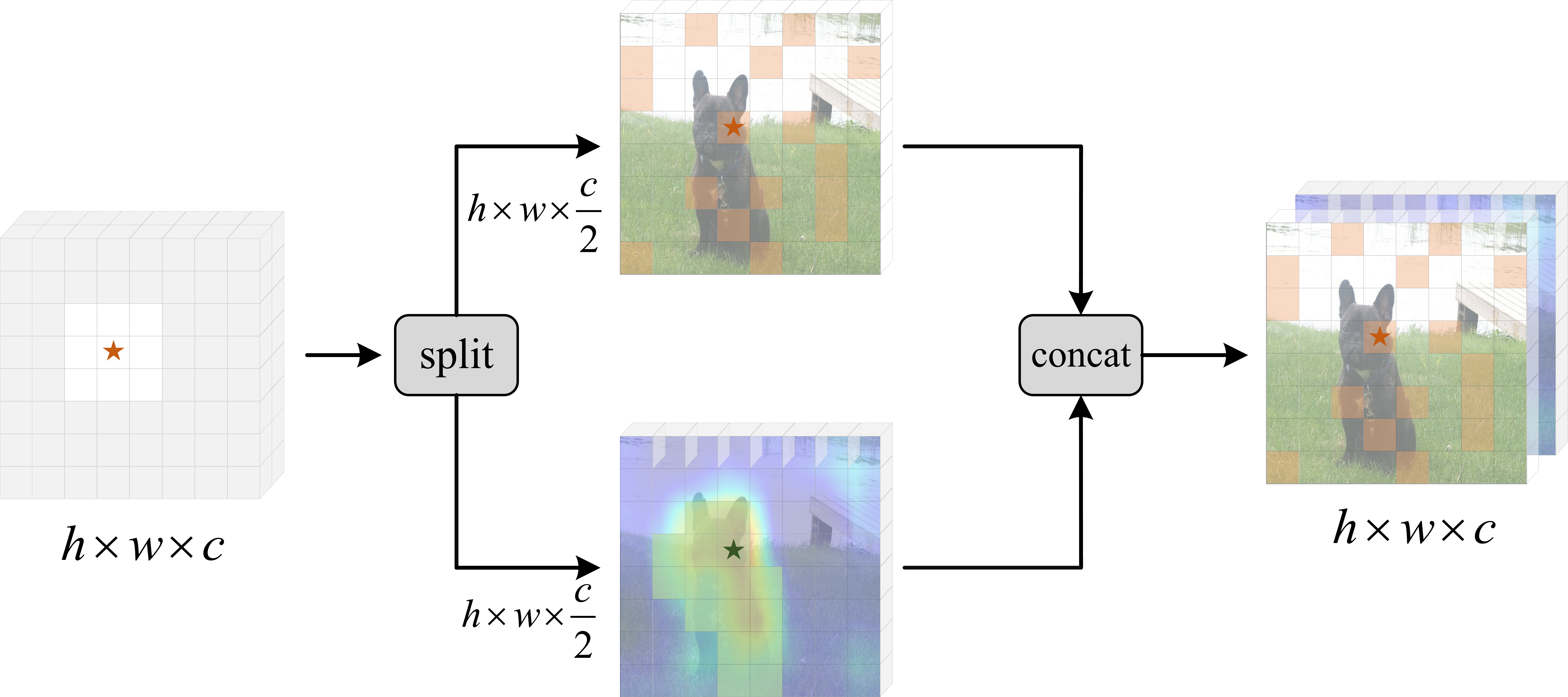}
   \caption{A parallel implementation of RSIR-Win.}
   \label{SplitGroup-flabel}
\end{figure}

\subsection{Parallel Implementation}

We split the $K$ heads into two parallel groups, $K/2$ heads per group, thus incorporating two different attention mechanisms, as shown in Figure \ref{SplitGroup-flabel}. The first group of heads perform RS-Win attention, the second group of heads perform IR-Win attention. Finally, the outputs of these two parallel groups will be concatenated back together.

\begin{equation}
\text{hea}{{\text{d}}_{k}}=\left\{ \begin{matrix}
   \text{RS-MS}{{\text{A}}_{k}}\text{(}X\text{)   }  \\
   \text{IR-MS}{{\text{A}}_{k}}\text{(}X\text{)   }  \\
\end{matrix}\begin{array}{*{35}{l}}
   k=1,\cdots ,K/2  \\
   k=K/2+1,\cdots ,K  \\
\end{array} \right.
  \label{mergingtoken-glabel}
\end{equation}

\begin{equation}
\text{RSIR-Win(}X\text{)=Concat(hea}{{\text{d}}_{1}}\text{,}\cdots \text{,hea}{{\text{d}}_{K}}\text{)}{{W}^{O}}
  \label{finalOutMlp-glabel}
\end{equation}
where ${{W}^{O}}\in {{R}^{C\times C}}$ is the commonly used projection matrix that is used to integrate the output tokens of two groups. Compared to the step-by-step implementation of RS-Win and IR-Win self-attention separately, such a parallel mechanism allows the incorporation of RS-Win and IR-Win information in each block and has a lower computation complexity.

\subsection{RSIR Transformer Block}
Equipped with the above self-attention mechanism, RSIR Transformer block is formally defined as:
\begin{equation}
\begin{aligned}
 & \overset{\wedge }{\mathop{{{X}^{l}}}}\,=\text{RSIR-Win}(\text{LN}({{X}^{l-1}}))+{{X}^{l-1}}, \\ 
 & {{X}^{l}}=\text{MLP}(\text{LN}(\overset{\wedge }{\mathop{{{X}^{l}}}}\,))+\overset{\wedge }{\mathop{{{X}^{l}}}}\, \\ 
\end{aligned}
  \label{windowAttention-glabel}
\end{equation}
where $\overset{\wedge }{\mathop{{{X}^{l}}}}\,$ and ${{X}^{l}}$ denote the output features of the $\text{RSIR-Win}$ module and the $\text{MLP}$ module for block $l$, respectively.

\begin{table*}[t]
   \centering
   \caption{Object detection and instance segmentation performance on the COCO val2017 with the Mask R-CNN framework and 1x training schedule. The models have been pre-trained on ImageNet-1K. The resolution used to calculate FLOPs is 800×1280.}
   \begin{tabular}{l|cc|cccccc}
      \hline % \toprule
      Backbone  & Params & FLOPs & $\text{AP}^{\text{box}}$ &  $\text{AP}^{\text{box}}_{\text{50}}$  &  $\text{AP}^{\text{box}}_{\text{75}}$ &  $\text{AP}^{\text{mask}}$ &  $\text{AP}^{\text{mask}}_{\text{50}}$ &  $\text{AP}^{\text{mask}}_{\text{75}}$          \\
      \hline % \midrule
      ResNet-50 \cite{He-2016-CVPR} & 44M   & 260G   &38.0  &58.6 &41.4&34.4&55.1&36.7     \\
      Twins-S \cite{NEURIPS2021-4e0928de} & 44M   & 228G   &42.7&65.6 &46.7&39.6&62.5&42.6    \\
      PVT-S \cite{Wang-2021-ICCV} & 44M   & 245G   &40.4&62.9 &43.8&37.8&60.1&40.3     \\
      Swin-T \cite{Liu-2021-ICCV} & 48M   & 264G   &43.7&66.6 &47.6&39.8&63.3&42.7    \\
      CSWin-T \cite{Dong-2022-CVPR} & 42M   & 279G   &46.7&68.6 &51.3&42.2&65.6&45.4     \\
      MaxViT-T \cite{10.1007/978-3-031-20053-3-27} & 49M   & 286G   &46.9&69.1 &52.0&42.2&66.1&46.3     \\
      RSIR-T (ours) & 42M   & 279G   &\textbf{47.6}&\textbf{69.2} &\textbf{52.3}&\textbf{42.7}&\textbf{66.3}&\textbf{46.6}          \\
      \hline % \midrule
      RegNeXt-101-64 \cite{He-2016-CVPR} & 101M   & 493G   &42.8  &63.8 &47.3&38.4&60.6&41.3     \\
      Twins-L \cite{NEURIPS2021-4e0928de} & 120M   & 474G   &45.2&67.5 &49.4&41.2&64.5&44.5    \\
      PVT-L \cite{Wang-2021-ICCV} & 81M   & 364G   &42.9&65.0 &46.6&39.5&61.9&42.5     \\
      Swin-B \cite{Liu-2021-ICCV} & 107M   & 496G   &46.9&-- &--&42.3&--&--      \\
      CSWin-B \cite{Dong-2022-CVPR} & 97M   & 526G   &48.7&70.4 &53.9&43.9&67.8&47.3     \\
      MaxViT-B \cite{10.1007/978-3-031-20053-3-27} & 103M   & 589G   &48.9&70.5 &54.1&43.8&67.9&47.6     \\
      RSIR-B (ours) & 96M & 517G &\textbf{49.2}&\textbf{71.0} &\textbf{54.1}&\textbf{44.5}&\textbf{68.1}&\textbf{47.7} \\
      \hline % \bottomrule
   \end{tabular}
   \label{COCO-Top1}
\end{table*}

\subsection{Overall Architecture} 
An overview architecture of the RSIR-ViT is presented in Figure \ref{OverallArchitecture-flabel} (a), which illustrates the tiny version. RSIR-ViT consists of four hierarchical stages, like Swin-ViT \cite{Liu-2021-ICCV} to build hierarchical architecture to capture multi-scale features. Each stage contains a patch merging layer and several RSIR Transformer blocks. As the network gets deeper, the input features are spatially downsampled by a certain ratio through the patch merging layer. The channel dimension is expanded twice to produce a hierarchical image representation. Specifically, the spatial downsampling ratio is set to 4 in the first stage and 2 in the last three stages. The outputs of the patch merging layer are fed into the subsequent RSIR Transformer block, and the number of tokens is kept constant. Finally, we apply a global average pooling step on the output of the last block to obtain the image representation vector for the final prediction.

\begin{table}[h]
   \centering
   \caption{Detailed conﬁgurations of RSIR Transformer Variants.}
   \resizebox{\linewidth}{!}{
   \begin{tabular}{l|c|c|c|c}
      \hline % \toprule
      Models  & \#Dim & \#Blocks  &\#heads   &\#Param.  \\
      \hline % \midrule
      RSIR-T &64   & 2,2,16,2   & 4,4,8,16       &22M     \\
      RSIR-B &96   & 2,4,24,2   & 4,8,16,32     &85M     \\
      \hline % \bottomrule
   \end{tabular}
   }
   \label{Variants-table}
\end{table}

\noindent \textbf{Variants.} For a fair comparison with other vision Transformers under similar settings, we designed two variants of the proposed RSIR Transformer: RSIR-T (Tiny) and RSIR-B (Base). Table \ref{Variants-table} shows the detailed conﬁgurations of all variants. They are designed by changing the block number of each stage and the base channel dimension $C$. The RSIR-T's head numbers for the four stages are 4, 4, 8, 16; the RSIR-B's head numbers are 4, 8, 16, 32.

\section{Experiments}

To show the effectiveness of the RSIR Transformer, we conduct experiments on ImageNet-1K \cite{5206848}. We then compare the performance of RSIR and state-of-the-art Transformer backbones on small datasets Caltech-256 \cite{griffin2007caltech} and Mini-ImageNet \cite{krizhevsky2012imagenet}. To further demonstrate the effectiveness and generalization of our backbone, we conduct experiments on ADE20K \cite{Zhou-2017-CVPR} for semantic segmentation, and COCO \cite{10.1007/978-3-319-10602-1-48} for object detection. Finally, we perform comprehensive ablation studies to analyze each component of the RSIR Transformer.

\subsection{Classiﬁcation on the ImageNet-1K}

\begin{table}[h]
   \centering
   \caption{Comparison of different models on ImageNet-1K.}
   \resizebox{\linewidth}{!}{
   \begin{tabular}{l|ccc|c}
      \hline % \toprule
      Method  & Image Size & Param. & FLOPs                & Top-1 acc.           \\
      \hline % \midrule
      RegNetY-4G \cite{Radosavovic-2020-CVPR} & ${{224}^{2}}$   & 21M   & 4.0G     &80.0    \\
      DeiT-S \cite{pmlr-v139-touvron21a} & ${{224}^{2}}$   & 22M   & 4.6G     &79.8    \\
      PVT-S \cite{Wang-2021-ICCV} & ${{224}^{2}}$   & 25M   & 3.8G     &79.8    \\
      Swin-T \cite{Liu-2021-ICCV} & ${{224}^{2}}$   & 29M   & 4.5G     &81.3    \\
      CSWin-T \cite{Dong-2022-CVPR} & ${{224}^{2}}$   & 23M   & 4.3G     &82.7     \\
      MaxViT-T \cite{10.1007/978-3-031-20053-3-27} & ${{224}^{2}}$   & 29M   & 5.6G     &83.6    \\
      RSIR-T (ours) & ${{224}^{2}}$   & 22M   & 5.0G     & \textbf{83.9}     \\
      \hline % \midrule
      RegNetY-16G \cite{Radosavovic-2020-CVPR} & ${{224}^{2}}$   & 84M   & 16.0G     &82.9    \\
      ViT-B  \cite{DBLP:journals/corr/abs-2010-11929} & ${{384}^{2}}$   & 86M   & 55.4G     &77.9     \\
      DeiT-B \cite{pmlr-v139-touvron21a} & ${{224}^{2}}$   & 86M   & 17.5G     &81.8    \\
      PVT-B \cite{Wang-2021-ICCV} & ${{224}^{2}}$   & 61M   & 9.8G     &81.7    \\
      Swin-B \cite{Liu-2021-ICCV} & ${{224}^{2}}$   & 88M   & 15.4G     &83.3     \\
      CSWin-B \cite{Dong-2022-CVPR} & ${{224}^{2}}$   & 78M   & 15.0G     &84.2     \\
      MaxViT-B \cite{10.1007/978-3-031-20053-3-27} & ${{224}^{2}}$   & 120M   & 23.4G     &84.9    \\
      RSIR-B (ours) & ${{224}^{2}}$   & 85M   & 16.3G     &\textbf{85.1}     \\
      \hline % \bottomrule
   \end{tabular}
   }
   \label{ImageNet-Top1}
\end{table}

\noindent \textbf{Implementation details.} This setting mostly follows \cite{Liu-2021-ICCV}. We use the PyTorch toolbox \cite{paszke2019pytorch} to implement all our experiments. We employ an AdamW \cite{kingma2014adam} optimizer for 300 epochs using a cosine decay learning rate scheduler and 20 epochs of linear warm-up. A batch size of 256, an initial learning rate of 0.001, and a weight decay of 0.05 are used. ViT-B/16 uses an image size 384×384 and others use 224×224. We include most of the augmentation and regularization strategies of Swin transformer\cite{Liu-2021-ICCV} in training.

\noindent \textbf{Results.} Table \ref{ImageNet-Top1} compares the performance of the proposed RSIR Transformer with the state-of-the-art CNN and Vision Transformer backbones on ImageNet-1K. Compared to ViT-B, the proposed RSIR-B model is +7.2\% better and has much lower computation complexity than ViT-B. Meanwhile, the proposed RSIR Transformer variants outperform the state-of-the-art Transformer-based backbones, and is +0.3\% higher than the most related MaxViT Transformer. RSIR Transformer has the low computation complexity compared to all models in Table \ref{ImageNet-Top1}. For example, RSIR-T achieves 83.9\% Top-1 accuracy with only 5.0G FLOPs. And for the base model setting, our RSIR-B also achieves the best performance.

\begin{table}[h]
   \centering
   \caption{Comparison of different models on Caltech-256.}
   \resizebox{\linewidth}{!}{
   \begin{tabular}{l|ccc|c}
      \hline % \toprule
      Method  & Image Size & Param. & FLOPs                & Top-1 acc.           \\
      \hline % \midrule
       Swin-T \cite{Liu-2021-ICCV} & ${{224}^{2}}$   & 29M   & 4.5G     &43.3    \\
       CSWin-T \cite{Dong-2022-CVPR} & ${{224}^{2}}$   & 23M   & 4.3G     &47.7     \\
       MaxViT-T \cite{10.1007/978-3-031-20053-3-27} & ${{224}^{2}}$   & 29M   & 5.6G     &47.9   \\
       RSIR-T (ours) & ${{224}^{2}}$   & 22M   & 5.0G     & \textbf{48.8}     \\
      \hline % \midrule
       ViT-B  \cite{DBLP:journals/corr/abs-2010-11929} & ${{384}^{2}}$   & 86M   & 55.4G     &37.6     \\
      Swin-B \cite{Liu-2021-ICCV} & ${{224}^{2}}$   & 88M   & 15.4G     &46.7     \\
      CSWin-B \cite{Dong-2022-CVPR} & ${{224}^{2}}$   & 78M   & 15.0G     &48.5     \\
      MaxViT-B \cite{10.1007/978-3-031-20053-3-27} & ${{224}^{2}}$   & 120M   & 23.4G     &48.6    \\
      RSIR-B (ours) & ${{224}^{2}}$   & 85M   & 16.3G     &\textbf{49.2}     \\
      \hline % \bottomrule
   \end{tabular}
   }
   \label{Caltech-256-Top1}
\end{table}

\subsection{Classiﬁcation on Caltech-256 and Mini-ImageNet}

\noindent \textbf{Implementation details.} Follow most of the experimental settings in the above subsection and change epochs to 100.

\noindent \textbf{Results.} In Table \ref{Caltech-256-Top1} and Table \ref{Mini-ImageNet-Top1}, we compare the proposed RSIR Transformer with state-of-the-art Transformer architectures on small datasets. With the limitation of pages, we only compare with a few classical methods here. It is known that ViTs usually perform poorly on such tasks as they typically require large datasets to be trained on. The models that perform well on large-scale ImageNet do not necessarily work perform on small-scale Mini-ImageNet and Caltech-256, e.g., ViT-B has top-1 accuracy of 58.3\% and Swin-B has top-1 accuracy of 67.4\% on the Mini-ImageNet, which suggests that ViTs are more challenging to train with less data. The proposed RSIR can significantly improve the data efficiency and performs well on small datasets such as Caltech-256 and Mini-ImageNet. Compared with MaxViT, it has increased by 0.9\% and 0.3\% respectively.

\begin{table}[h]
   \centering
   \caption{Comparison of different models on Mini-ImageNet.}
   \resizebox{\linewidth}{!}{
   \begin{tabular}{l|ccc|c}
      \hline % \toprule
      Method  & Image Size & Param. & FLOPs                & Top-1 acc.           \\
      \hline % \midrule
      Swin-T \cite{Liu-2021-ICCV} & ${{224}^{2}}$   & 29M   & 4.5G     &66.3    \\
      CSWin-T \cite{Dong-2022-CVPR} & ${{224}^{2}}$   & 23M   & 4.3G     &66.8     \\
      MaxViT-T \cite{10.1007/978-3-031-20053-3-27} & ${{224}^{2}}$   & 29M   & 5.6G     &67.9   \\
      RSIR-T (ours) & ${{224}^{2}}$   & 22M   & 5.0G     & \textbf{68.4}     \\
      \hline % \midrule
      ViT-B  \cite{DBLP:journals/corr/abs-2010-11929} & ${{384}^{2}}$   & 86M   & 55.4G     &58.3     \\
      Swin-B \cite{Liu-2021-ICCV} & ${{224}^{2}}$   & 88M   & 15.4G     &67.4     \\
      CSWin-B \cite{Dong-2022-CVPR} & ${{224}^{2}}$   & 78M   & 15.0G     &68.4     \\
      MaxViT-B \cite{10.1007/978-3-031-20053-3-27} & ${{224}^{2}}$   & 120M   & 23.4G     &68.6    \\
      RSIR-B (ours) & ${{224}^{2}}$   & 85M   & 16.3G     &\textbf{69.1}     \\
      \hline % \bottomrule
   \end{tabular}
   }
   \label{Mini-ImageNet-Top1}
\end{table}

\subsection{COCO Object Detection}

\noindent \textbf{Implementation details.} We use the Mask R-CNN \cite{He-2017-ICCV} framework to evaluate the performance of the proposed RSIR Transformer backbone on the COCO benchmark for object detection. We pretrain the backbones on the ImageNet-1K dataset and apply the ﬁnetuning strategy used in Swin Transformer \cite{Liu-2021-ICCV} on the COCO training set.

\noindent \textbf{Results.} We compare RSIR Transformer with various backbones, as shown in Table \ref{COCO-Top1}. It shows that the proposed RSIR Transformer variants clearly outperform all the CNN and Transformer counterparts. For object detection, our RSIR-T and RSIR-B achieve 47.6 and 49.2 box mAP for object detection, surpassing the previous best CSWin Transformer by +0.7 and +0.3, respectively. We also achieve similar performance gain on instance segmentation.

\begin{table}[t]
   \centering
   \caption{Comparison of the segmentation performance of different backbones on the ADE20K. All backbones are pretrained on ImageNet-1K with the size of 224 ×224. The resolution used to calculate FLOPs is 512 ×2048.}
   \resizebox{\linewidth}{!}{
   \begin{tabular}{l|cc|cc}
      \hline % \toprule
      Backbone  & Params & FLOPs & SS mIoU & MS mIoU     \\
      \hline % \midrule
      Twins-S \cite{NEURIPS2021-4e0928de} & 55M   & 905G   &46.2&47.1     \\
      Swin-T \cite{Liu-2021-ICCV} & 60M   & 945G   &44.5&45.8    \\
      CSWin-T \cite{Dong-2022-CVPR} & 60M   & 959G   &49.3&50.4     \\
      MaxViT-T \cite{10.1007/978-3-031-20053-3-27} & 71M   & 989G   &49.5&50.7     \\
      RSIR-T (ours) & 57M   & 956G   &\textbf{49.9}&\textbf{51.2}           \\
      \hline % \midrule
      Twins-L \cite{NEURIPS2021-4e0928de} & 113M   & 1164G   &48.8&50.2     \\
      Swin-B \cite{Liu-2021-ICCV} & 121M   & 1188G   &48.1&49.7     \\
      CSWin-B \cite{Dong-2022-CVPR} & 109M   & 1222G   &50.8&51.7     \\
      MaxViT-B \cite{10.1007/978-3-031-20053-3-27} & 125M   & 1221G   &51.3&52.6     \\
      RSIR-B (ours) & 109M & 1210G &\textbf{51.9}&\textbf{52.7}  \\
      \hline % \bottomrule
   \end{tabular}
   \label{ADE20K-Top1}
   }
\end{table}

\subsection{ADE20K Semantic Segmentation}

\noindent \textbf{Implementation details.} We further investigate the capability of RSIR Transformer for Semantic Segmentation on the ADE20K \cite{Zhou-2017-CVPR} dataset. Here we employ the widely-used UperNet \cite{Xiao-2018-ECCV} as the basic framework and followed Swin's \cite{Liu-2021-ICCV} experimental settings In Table \ref{ADE20K-Top1}, we report both the single-scale (SS) and multi-scale (MS) mIoU for better comparison.

\noindent \textbf{Results.} As shown in Table \ref{ADE20K-Top1}, our RSIR variants outperform previous state-of-the-arts under different conﬁgurations. Speciﬁcally, our RSIR-T and RSIR-B outperform the MaxViT by +0.4\% and +0.6\% SS mIoU, respectively. These results show that the proposed RSIR Transformer can effectively capture the context dependencies of different distances.

\subsection{Ablation Study}

We perform ablation studies on image classification and downstream tasks for the fundamental designs of our RSIR Transformer. For a fair comparison, we use the Swin-T \cite{Liu-2021-ICCV} as the backbone for the following experiments, and only change one component for each ablation.

\noindent \textbf{Attention Mechanism Comparison.} In this subsection, we compare with existing self-attention mechanisms. As shown in Table \ref{differentMechanisms}, the proposed RSIR self-attention mechanism performs better than the existing self-attention mechanism.

\begin{table}[h]
   \centering
   \caption{Comparison of  different self-attention  mechanisms.}
   \resizebox{\linewidth}{!}{
   \begin{tabular}{l|ccc}
      \hline % \toprule
        & ImageNet & COCO & ADE20k                    \\
         & top-1   & $\text{AP}^{\text{box}}$    &SS mIoU     \\
      \hline % \midrule
       Swin's shifted windows \cite{Liu-2021-ICCV}   & 81.3    &43.7  &44.5    \\
       Spatially Sep \cite{NEURIPS2021-4e0928de}   & 81.5    &44.2  &45.8    \\
       Sequential Axial \cite{DBLP:journals/corr/abs-1912-12180}   & 81.5    &41.5  &42.9    \\
       Criss-Cross \cite{Huang-2019-ICCV}   & 81.7    &44.5  &45.9    \\
       Cross-shaped  \cite{Dong-2022-CVPR}   & 82.2    &45.0  &46.2    \\
       Grid \cite{10.1007/978-3-031-20053-3-27}   & 82.5    &45.2  &46.2    \\
       RSIR (ours)   & \textbf{82.9}    &\textbf{45.5}  &\textbf{46.3}    \\
      \hline % \bottomrule
   \end{tabular}
   }
   \label{differentMechanisms}
\end{table}

\section{Conclusions}

In this paper, we have presented a new Vision Transformer architecture named RSIR-Win Transformer. The core design of RSIR-Win Transformer consists of two components: random sampling windows (RS-Win) and important region windows (IR-Win). RS-Win can sample patches anywhere in the image, increasing the flexibility of the model's global information flow. IR-Win uses an attention map of input features for sampling, and the composed window is data-dependent. On the other hand, the RSIR-Win Transformer performs RS-Win and IR-Win self-attention in parallel by splitting the multi-heads into two parallel groups. This multi-head grouping design allows the model to efficiently incorporate information from both components without extra computation cost. RSIR-Win Transformer can achieve state-of-the-art performance on ImageNet-1K image classification, COCO object detection and ADE20K semantic segmentation. 

\bibliographystyle{named}
\bibliography{ijcai22}

\begin{thebibliography}{}

\bibitem[\protect\citeauthoryear{Chu \bgroup \em et al.\egroup
  }{2021a}]{NEURIPS2021-4e0928de}
Xiangxiang Chu, Zhi Tian, Yuqing Wang, Bo~Zhang, Haibing Ren, Xiaolin Wei,
  Huaxia Xia, and Chunhua Shen.
\newblock Twins: Revisiting the design of spatial attention in vision
  transformers.
\newblock In M.~Ranzato, A.~Beygelzimer, Y.~Dauphin, P.S. Liang, and J.~Wortman
  Vaughan, editors, {\em Advances in Neural Information Processing Systems},
  volume~34, pages 9355--9366. Curran Associates, Inc., 2021.

\bibitem[\protect\citeauthoryear{Chu \bgroup \em et al.\egroup
  }{2021b}]{DBLP:journals/corr/abs-2102-10882}
Xiangxiang Chu, Bo~Zhang, Zhi Tian, Xiaolin Wei, and Huaxia Xia.
\newblock Do we really need explicit position encodings for vision
  transformers?
\newblock {\em CoRR}, abs/2102.10882, 2021.

\bibitem[\protect\citeauthoryear{Dai \bgroup \em et al.\egroup
  }{2021}]{NEURIPS2021-20568692}
Zihang Dai, Hanxiao Liu, Quoc~V Le, and Mingxing Tan.
\newblock Coatnet: Marrying convolution and attention for all data sizes.
\newblock In M.~Ranzato, A.~Beygelzimer, Y.~Dauphin, P.S. Liang, and J.~Wortman
  Vaughan, editors, {\em Advances in Neural Information Processing Systems},
  volume~34, pages 3965--3977. Curran Associates, Inc., 2021.

\bibitem[\protect\citeauthoryear{Deng \bgroup \em et al.\egroup
  }{2009}]{5206848}
Jia Deng, Wei Dong, Richard Socher, Li-Jia Li, Kai Li, and Li~Fei-Fei.
\newblock Imagenet: A large-scale hierarchical image database.
\newblock In {\em 2009 IEEE Conference on Computer Vision and Pattern
  Recognition}, pages 248--255, 2009.

\bibitem[\protect\citeauthoryear{Devlin \bgroup \em et al.\egroup
  }{2018}]{DBLP:journals/corr/abs-1810-04805}
Jacob Devlin, Ming{-}Wei Chang, Kenton Lee, and Kristina Toutanova.
\newblock {BERT:} pre-training of deep bidirectional transformers for language
  understanding.
\newblock {\em CoRR}, abs/1810.04805, 2018.

\bibitem[\protect\citeauthoryear{Dong \bgroup \em et al.\egroup
  }{2022}]{Dong-2022-CVPR}
Xiaoyi Dong, Jianmin Bao, Dongdong Chen, Weiming Zhang, Nenghai Yu, Lu~Yuan,
  Dong Chen, and Baining Guo.
\newblock Cswin transformer: A general vision transformer backbone with
  cross-shaped windows.
\newblock In {\em Proceedings of the IEEE/CVF Conference on Computer Vision and
  Pattern Recognition (CVPR)}, pages 12124--12134, June 2022.

\bibitem[\protect\citeauthoryear{Dosovitskiy \bgroup \em et al.\egroup
  }{2020}]{DBLP:journals/corr/abs-2010-11929}
Alexey Dosovitskiy, Lucas Beyer, Alexander Kolesnikov, Dirk Weissenborn,
  Xiaohua Zhai, Thomas Unterthiner, Mostafa Dehghani, Matthias Minderer, Georg
  Heigold, Sylvain Gelly, Jakob Uszkoreit, and Neil Houlsby.
\newblock An image is worth 16x16 words: Transformers for image recognition at
  scale.
\newblock {\em CoRR}, abs/2010.11929, 2020.

\bibitem[\protect\citeauthoryear{Griffin \bgroup \em et al.\egroup
  }{2007}]{griffin2007caltech}
Gregory Griffin, Alex Holub, and Pietro Perona.
\newblock Caltech-256 object category dataset.
\newblock 2007.

\bibitem[\protect\citeauthoryear{He \bgroup \em et al.\egroup
  }{2016}]{He-2016-CVPR}
Kaiming He, Xiangyu Zhang, Shaoqing Ren, and Jian Sun.
\newblock Deep residual learning for image recognition.
\newblock In {\em Proceedings of the IEEE Conference on Computer Vision and
  Pattern Recognition (CVPR)}, June 2016.

\bibitem[\protect\citeauthoryear{He \bgroup \em et al.\egroup
  }{2017}]{He-2017-ICCV}
Kaiming He, Georgia Gkioxari, Piotr Dollar, and Ross Girshick.
\newblock Mask r-cnn.
\newblock In {\em Proceedings of the IEEE International Conference on Computer
  Vision (ICCV)}, Oct 2017.

\bibitem[\protect\citeauthoryear{Ho \bgroup \em et al.\egroup
  }{2019}]{DBLP:journals/corr/abs-1912-12180}
Jonathan Ho, Nal Kalchbrenner, Dirk Weissenborn, and Tim Salimans.
\newblock Axial attention in multidimensional transformers.
\newblock {\em CoRR}, abs/1912.12180, 2019.

\bibitem[\protect\citeauthoryear{Huang \bgroup \em et al.\egroup
  }{2018}]{Huang-2018-CVPR}
Gao Huang, Shichen Liu, Laurens van~der Maaten, and Kilian~Q. Weinberger.
\newblock Condensenet: An efficient densenet using learned group convolutions.
\newblock In {\em Proceedings of the IEEE Conference on Computer Vision and
  Pattern Recognition (CVPR)}, June 2018.

\bibitem[\protect\citeauthoryear{Huang \bgroup \em et al.\egroup
  }{2019}]{Huang-2019-ICCV}
Zilong Huang, Xinggang Wang, Lichao Huang, Chang Huang, Yunchao Wei, and Wenyu
  Liu.
\newblock Ccnet: Criss-cross attention for semantic segmentation.
\newblock In {\em Proceedings of the IEEE/CVF International Conference on
  Computer Vision (ICCV)}, October 2019.

\bibitem[\protect\citeauthoryear{Kingma and Ba}{2014}]{kingma2014adam}
Diederik~P Kingma and Jimmy Ba.
\newblock Adam: A method for stochastic optimization.
\newblock {\em arXiv preprint arXiv:1412.6980}, 2014.

\bibitem[\protect\citeauthoryear{Krizhevsky \bgroup \em et al.\egroup
  }{2012}]{krizhevsky2012imagenet}
Alex Krizhevsky, Ilya Sutskever, and Geoffrey~E Hinton.
\newblock Imagenet classification with deep convolutional neural networks.
\newblock {\em Advances in neural information processing systems},
  25:1097--1105, 2012.

\bibitem[\protect\citeauthoryear{Lin \bgroup \em et al.\egroup
  }{2014}]{10.1007/978-3-319-10602-1-48}
Tsung-Yi Lin, Michael Maire, Serge Belongie, James Hays, Pietro Perona, Deva
  Ramanan, Piotr Doll{\'a}r, and C.~Lawrence Zitnick.
\newblock Microsoft coco: Common objects in context.
\newblock In David Fleet, Tomas Pajdla, Bernt Schiele, and Tinne Tuytelaars,
  editors, {\em Computer Vision -- ECCV 2014}, pages 740--755, Cham, 2014.
  Springer International Publishing.

\bibitem[\protect\citeauthoryear{Liu \bgroup \em et al.\egroup
  }{2021}]{Liu-2021-ICCV}
Ze~Liu, Yutong Lin, Yue Cao, Han Hu, Yixuan Wei, Zheng Zhang, Stephen Lin, and
  Baining Guo.
\newblock Swin transformer: Hierarchical vision transformer using shifted
  windows.
\newblock In {\em Proceedings of the IEEE/CVF International Conference on
  Computer Vision (ICCV)}, pages 10012--10022, October 2021.

\bibitem[\protect\citeauthoryear{Paszke \bgroup \em et al.\egroup
  }{2019}]{paszke2019pytorch}
Adam Paszke, Sam Gross, Francisco Massa, Adam Lerer, James Bradbury, Gregory
  Chanan, Trevor Killeen, Zeming Lin, Natalia Gimelshein, Luca Antiga, et~al.
\newblock Pytorch: An imperative style, high-performance deep learning library.
\newblock {\em Advances in neural information processing systems},
  32:8026--8037, 2019.

\bibitem[\protect\citeauthoryear{Radosavovic \bgroup \em et al.\egroup
  }{2020}]{Radosavovic-2020-CVPR}
Ilija Radosavovic, Raj~Prateek Kosaraju, Ross Girshick, Kaiming He, and Piotr
  Dollar.
\newblock Designing network design spaces.
\newblock In {\em Proceedings of the IEEE/CVF Conference on Computer Vision and
  Pattern Recognition (CVPR)}, June 2020.

\bibitem[\protect\citeauthoryear{Touvron \bgroup \em et al.\egroup
  }{2021}]{pmlr-v139-touvron21a}
Hugo Touvron, Matthieu Cord, Matthijs Douze, Francisco Massa, Alexandre
  Sablayrolles, and Herve Jegou.
\newblock Training data-efficient image transformers-amp; distillation through
  attention.
\newblock In Marina Meila and Tong Zhang, editors, {\em Proceedings of the 38th
  International Conference on Machine Learning}, volume 139 of {\em Proceedings
  of Machine Learning Research}, pages 10347--10357. PMLR, 18--24 Jul 2021.

\bibitem[\protect\citeauthoryear{Tu \bgroup \em et al.\egroup
  }{2022}]{10.1007/978-3-031-20053-3-27}
Zhengzhong Tu, Hossein Talebi, Han Zhang, Feng Yang, Peyman Milanfar, Alan
  Bovik, and Yinxiao Li.
\newblock Maxvit: Multi-axis vision transformer.
\newblock In Shai Avidan, Gabriel Brostow, Moustapha Ciss{\'e}, Giovanni~Maria
  Farinella, and Tal Hassner, editors, {\em Computer Vision -- ECCV 2022},
  pages 459--479, Cham, 2022. Springer Nature Switzerland.

\bibitem[\protect\citeauthoryear{Vaswani \bgroup \em et al.\egroup
  }{2017}]{NIPS2017-3f5ee243}
Ashish Vaswani, Noam Shazeer, Niki Parmar, Jakob Uszkoreit, Llion Jones,
  Aidan~N Gomez, \L~ukasz Kaiser, and Illia Polosukhin.
\newblock Attention is all you need.
\newblock In I.~Guyon, U.~Von Luxburg, S.~Bengio, H.~Wallach, R.~Fergus,
  S.~Vishwanathan, and R.~Garnett, editors, {\em Advances in Neural Information
  Processing Systems}, volume~30. Curran Associates, Inc., 2017.

\bibitem[\protect\citeauthoryear{Vaswani \bgroup \em et al.\egroup
  }{2021}]{Vaswani-2021-CVPR}
Ashish Vaswani, Prajit Ramachandran, Aravind Srinivas, Niki Parmar, Blake
  Hechtman, and Jonathon Shlens.
\newblock Scaling local self-attention for parameter efficient visual
  backbones.
\newblock In {\em Proceedings of the IEEE/CVF Conference on Computer Vision and
  Pattern Recognition (CVPR)}, pages 12894--12904, June 2021.

\bibitem[\protect\citeauthoryear{Wang \bgroup \em et al.\egroup
  }{2021}]{Wang-2021-ICCV}
Wenhai Wang, Enze Xie, Xiang Li, Deng-Ping Fan, Kaitao Song, Ding Liang, Tong
  Lu, Ping Luo, and Ling Shao.
\newblock Pyramid vision transformer: A versatile backbone for dense prediction
  without convolutions.
\newblock In {\em Proceedings of the IEEE/CVF International Conference on
  Computer Vision (ICCV)}, pages 568--578, October 2021.

\bibitem[\protect\citeauthoryear{Woo \bgroup \em et al.\egroup
  }{2018}]{Woo-2018-ECCV}
Sanghyun Woo, Jongchan Park, Joon-Young Lee, and In~So Kweon.
\newblock Cbam: Convolutional block attention module.
\newblock In {\em Proceedings of the European Conference on Computer Vision
  (ECCV)}, September 2018.

\bibitem[\protect\citeauthoryear{Xiao \bgroup \em et al.\egroup
  }{2018}]{Xiao-2018-ECCV}
Tete Xiao, Yingcheng Liu, Bolei Zhou, Yuning Jiang, and Jian Sun.
\newblock Unified perceptual parsing for scene understanding.
\newblock In {\em Proceedings of the European Conference on Computer Vision
  (ECCV)}, September 2018.

\bibitem[\protect\citeauthoryear{Yang \bgroup \em et al.\egroup
  }{2021}]{DBLP:journals/corr/abs-2107-00641}
Jianwei Yang, Chunyuan Li, Pengchuan Zhang, Xiyang Dai, Bin Xiao, Lu~Yuan, and
  Jianfeng Gao.
\newblock Focal self-attention for local-global interactions in vision
  transformers.
\newblock {\em CoRR}, abs/2107.00641, 2021.

\bibitem[\protect\citeauthoryear{Yates \bgroup \em et al.\egroup
  }{2021}]{10.1145/3437963.3441667}
Andrew Yates, Rodrigo Nogueira, and Jimmy Lin.
\newblock Pretrained transformers for text ranking: Bert and beyond.
\newblock In {\em Proceedings of the 14th ACM International Conference on Web
  Search and Data Mining}, WSDM '21, page 1154–1156, New York, NY, USA, 2021.
  Association for Computing Machinery.

\bibitem[\protect\citeauthoryear{Zhang \bgroup \em et al.\egroup
  }{2022}]{zhang2022nested}
Zizhao Zhang, Han Zhang, Long Zhao, Ting Chen, Sercan~O Arik, and Tomas
  Pfister.
\newblock Nested hierarchical transformer: Towards accurate, data-efficient and
  interpretable visual understanding.
\newblock In {\em AAAI Conference on Artificial Intelligence (AAAI)}, volume
  2022, 2022.

\bibitem[\protect\citeauthoryear{Zhao \bgroup \em et al.\egroup
  }{2020}]{Zhao2020CVPR}
Hengshuang Zhao, Jiaya Jia, and Vladlen Koltun.
\newblock Exploring self-attention for image recognition.
\newblock In {\em Proceedings of the IEEE/CVF Conference on Computer Vision and
  Pattern Recognition (CVPR)}, June 2020.

\bibitem[\protect\citeauthoryear{Zhou \bgroup \em et al.\egroup
  }{2017}]{Zhou-2017-CVPR}
Bolei Zhou, Hang Zhao, Xavier Puig, Sanja Fidler, Adela Barriuso, and Antonio
  Torralba.
\newblock Scene parsing through ade20k dataset.
\newblock In {\em Proceedings of the IEEE Conference on Computer Vision and
  Pattern Recognition (CVPR)}, July 2017.

\end{thebibliography}

\end{document}